\documentclass{article} % For LaTeX2e
\usepackage{iclr2017_workshop,times}
\usepackage{url}
\usepackage{float}

\usepackage{amssymb}
\usepackage{amsmath}
\usepackage{subcaption}
\usepackage{graphicx}
\usepackage{microtype}

\title{Adversarial Examples for \\
Semantic Image Segmentation}

\author{%
Volker Fischer$^{1}$, Mummadi Chaithanya Kumar$^{2}$, Jan Hendrik Metzen$^{1}$ \& Thomas Brox$^{2}$ \\
$^{1}$ Bosch Center for Artificial Intelligence, Robert Bosch GmbH \\
$^{2}$ University of Freiburg \\
\texttt{\{volker.fischer, janhendrik.metzen\}@de.bosch.com} \\
\texttt{chaithu0536@gmail.com}; \texttt{brox@cs.uni-freiburg.de} \\
}

\begin{document}

\maketitle

\begin{abstract}
Machine learning methods in general and Deep Neural Networks in particular have shown to be vulnerable to adversarial perturbations. So far this phenomenon has mainly been studied in the context of whole-image classification. In this contribution, we analyse how adversarial perturbations can affect the task of semantic segmentation. We show how existing adversarial attackers can be transferred to this task and that it is possible to create imperceptible adversarial perturbations that lead a deep network to misclassify almost all pixels of a chosen class while leaving network prediction nearly unchanged outside this class.
\end{abstract}

\section{Introduction}
\label{Section:Introduction}

While machine learning methods and in particular deep neural networks have led to significant performance increases for numerous tasks,
several studies have found them to be vulnerable to adversarial attackers \citep{szegedy_intriguing_2013}. These attackers compute
perturbed versions of input data that fool the classifier to change its predictions on the new version while these perturbations stay almost imperceptible to the human eye. Most of the work on adversarial examples focusses on the task of image classification. In this paper, we investigate the effect of adversarial attacks on a localization task, in particular semantic segmentation.

Our work uses a common deep neural network for semantic segmentation and evaluates to which extent this task is vulnerable to adversarial examples. We introduce a way how adversarial examples could be defined for semantic segmentation and show how existing methods to create adversarial perturbations can be adapted for semantic segmentation. On the basis of the Cityscapes dataset, we find that it is easily possible to let a deep network misclassify entire regions of an image while replacing these misclassified areas with a convenient substitution. Finally we introduce a measure to analyse the effectiveness of a given attacker for a semantic segmentation network.

\section{Methods}
\label{section:Methods}

\paragraph{Generating adversarial examples}
\label{paragraph:Generating adversarial examples}

Let $f_{\theta}$ denote a function for semantic segmentation such as a deep neural network with parameters $\theta$ and $x$ an input image of size $W\times H$ with ground-truth label $y^{\text{true}}$. As most semantic segmentation tasks, we assume that $y^{\text{true}}$ has the same pixel dimensions as
the image $x$ times the number of classes. We also denote with $J_{\text{cls}}(f_{\theta}(x),y)$ the classification loss (i.e. cross-entropy) and assume it is differentiable with respect to $\theta$ and $x$.

Since adversarial examples were discovered by \citet{szegedy_intriguing_2013}, several methods have been introduced to generate adversarial perturbations (e.g. \citep{kurakin_adversarial_2016}, \citep{parsing_2015}, \citep{goodfellow_explaining_2015}). In the scope of this work we will focus on the so-called \emph{least likely} method presented by \citet{kurakin_adversarial_2016}. Despite its name this method is not restricted to target the least-likely class, but, in contrast to most other approaches, allows to explicitly specify a target class we want the classifier to predict. This method can easily be transferred from a single pixel to an entire image.

\begin{figure}[H]
\begin{center}
\begin{tabular}{ccc}
(a) original image & (b) adv. example & (c) adv. restricted \\
\includegraphics[width=.3\linewidth]{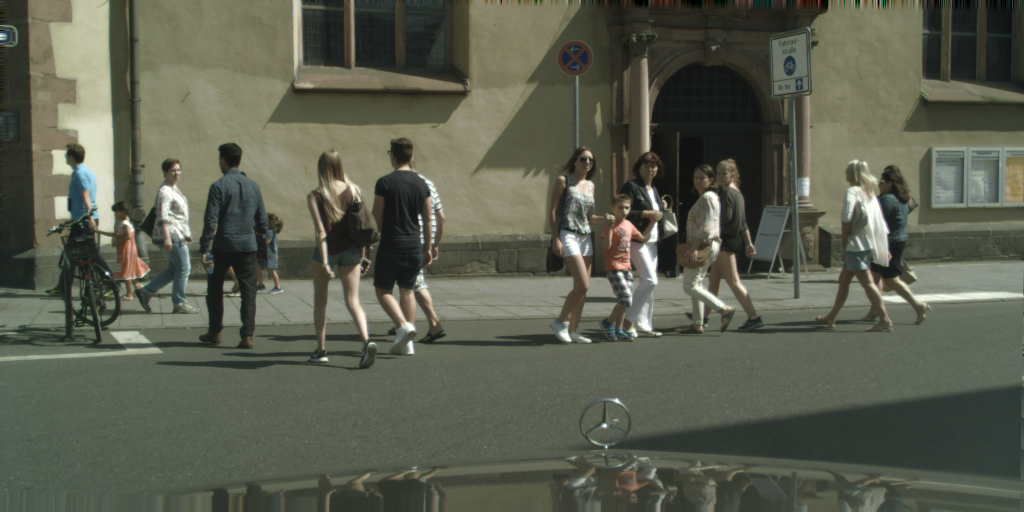} & 
\includegraphics[width=.3\linewidth]{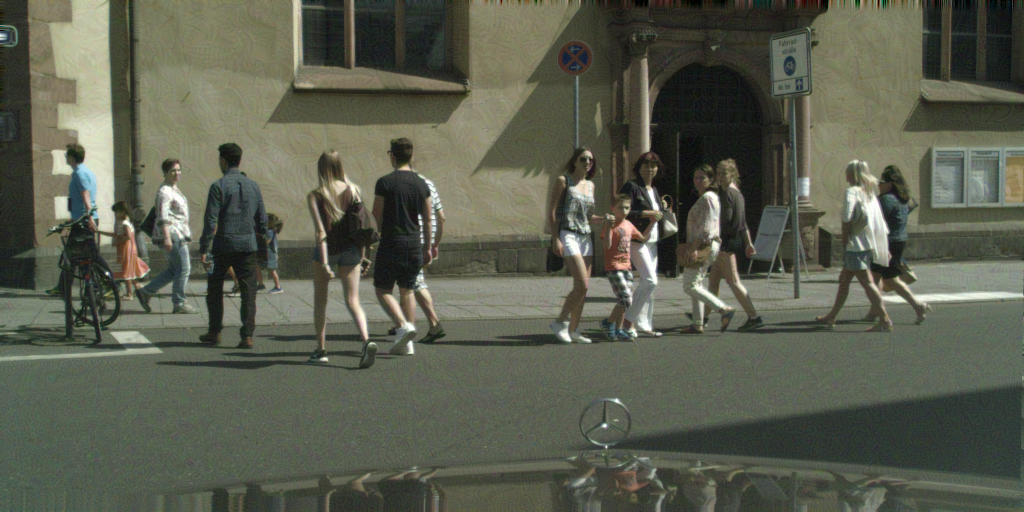} & 
\includegraphics[width=.3\linewidth]{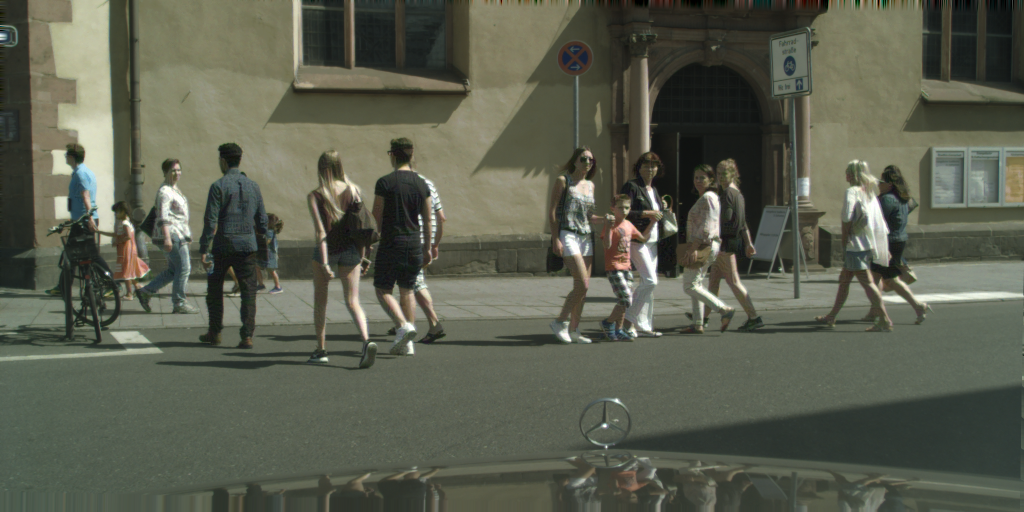} \\

 & (d) adv. noise ($\times 8$) & (e) adv. restricted noise ($\times 8$) \\
 &
\includegraphics[width=.3\linewidth]{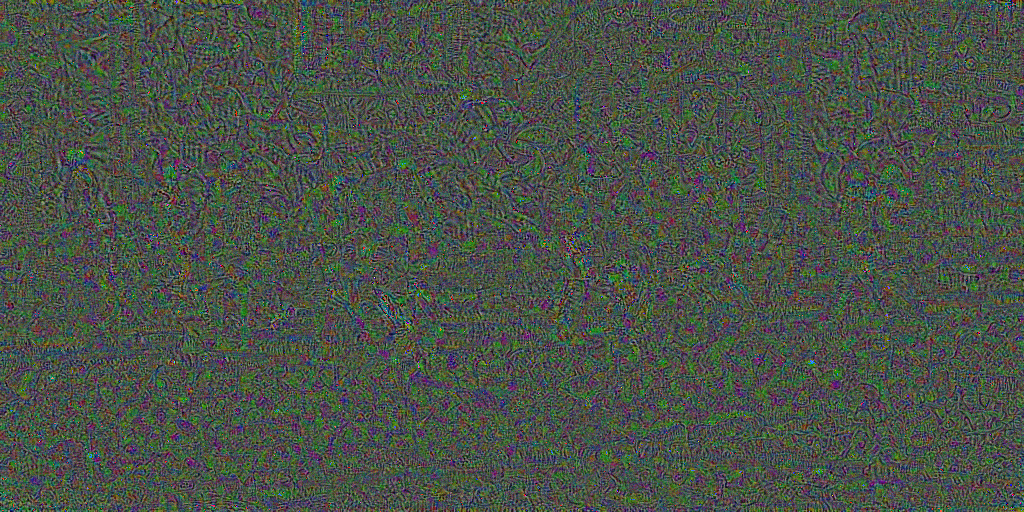} & 
\includegraphics[width=.3\linewidth]{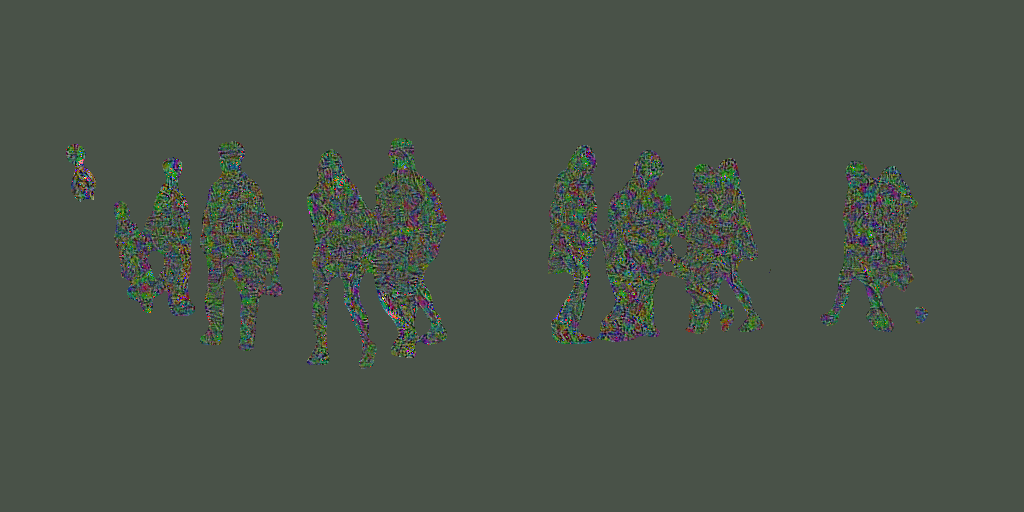} \\

(f) adv. target & (g) pred. on adv. & (h) pred on restricted adv. \\
\includegraphics[width=.3\linewidth]{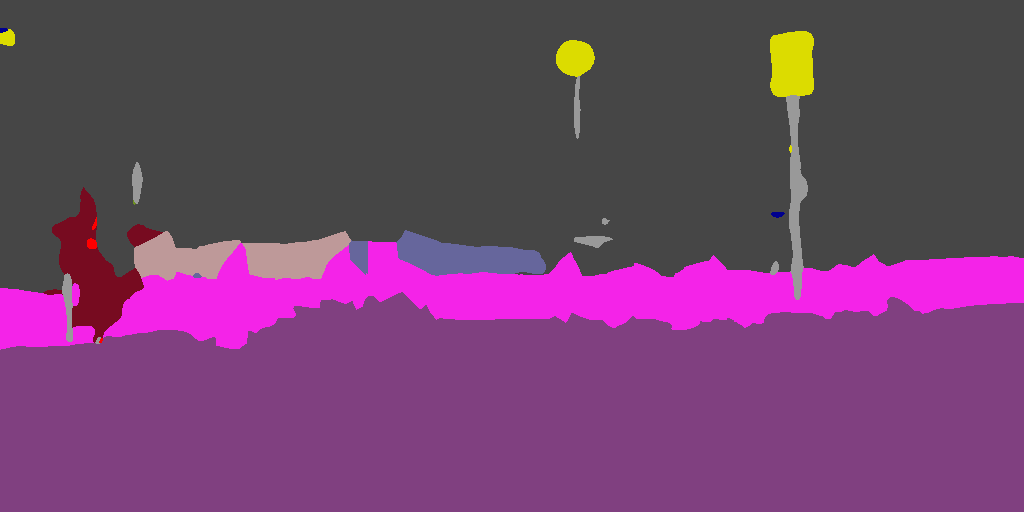} & 
\includegraphics[width=.3\linewidth]{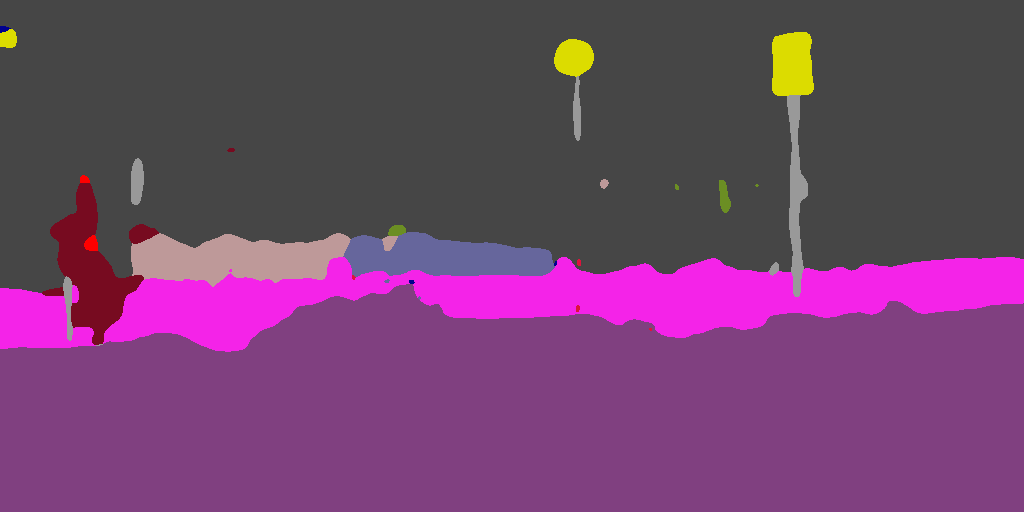} & 
\includegraphics[width=.3\linewidth]{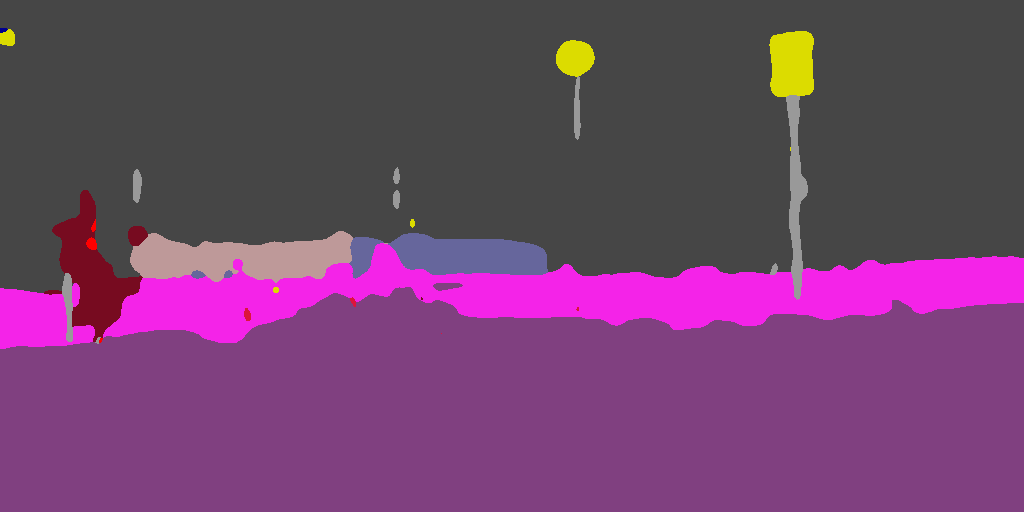} \\

(i) prediction & (j) pred. vs. pred. on adv. & (k) pred. vs. pred. rest. adv. \\
\includegraphics[width=.3\linewidth]{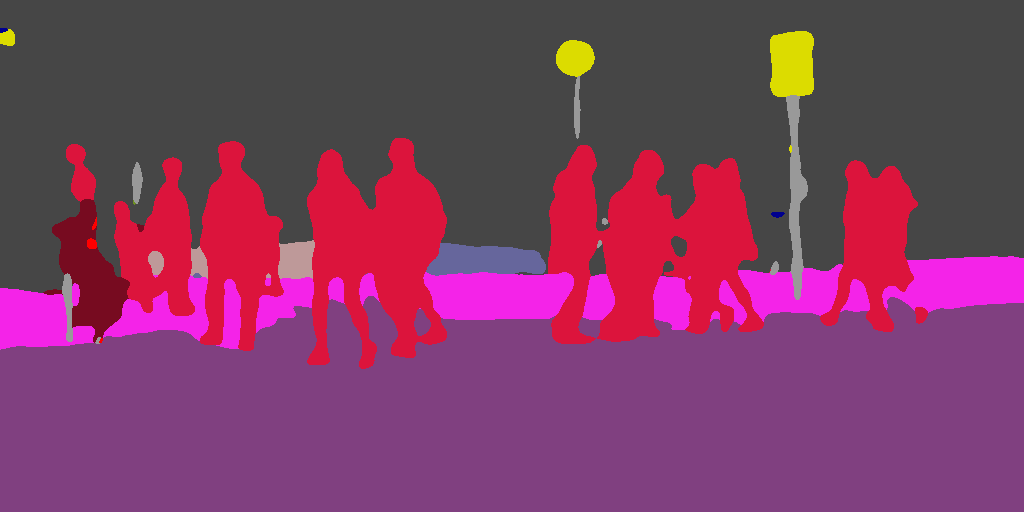} & 
\includegraphics[width=.3\linewidth]{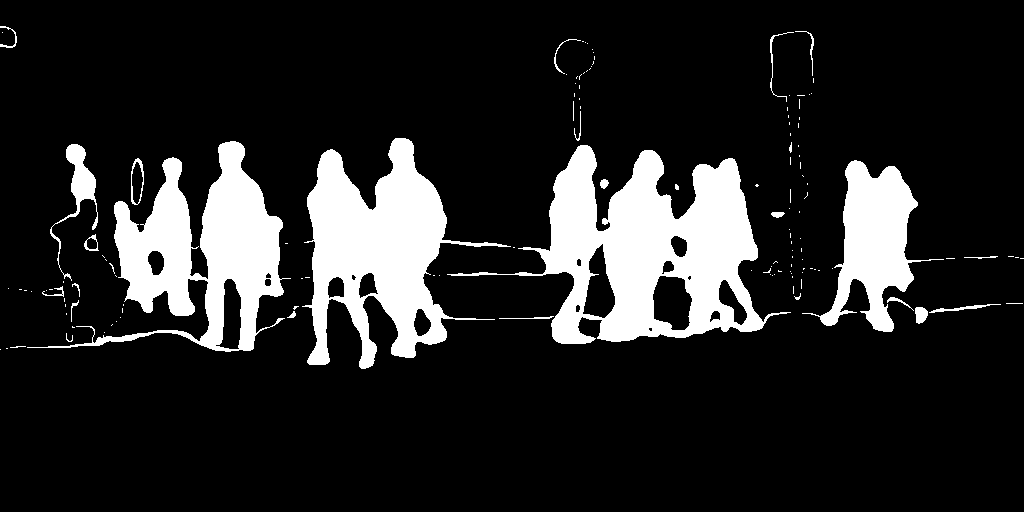} & 
\includegraphics[width=.3\linewidth]{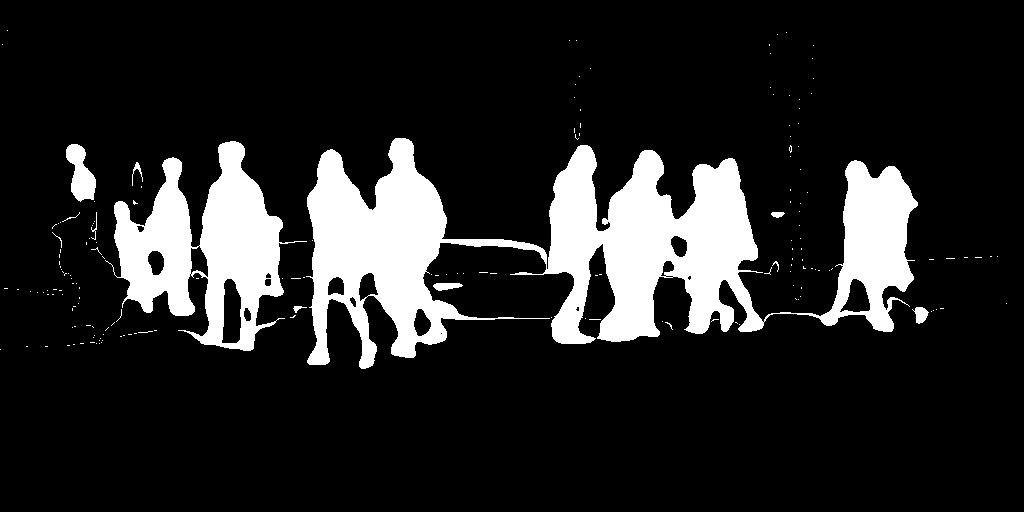} \\

%(k) adv. target & (l) target vs. pred. on adv. & (m) target vs. pred. rest. adv. \\
%\includegraphics[width=.3\linewidth]{images_for_paper/sample_images/frankfurt_000001_055172_/Adversarial_target.png} & 
%\includegraphics[width=.3\linewidth]{images_for_paper/sample_images/frankfurt_000001_055172_/differences_AdvTarget_AdvPred_eps_10.png} & 
%\includegraphics[width=.3\linewidth]{images_for_paper/sample_images/frankfurt_000001_055172_noise_limited/differences_AdvTarget_AdvPred_eps10.png}
\end{tabular}
\end{center}
\caption{Adversarial example for the task of semantic segmentation: \textbf{(a)} The original Cityscapes image. \textbf{(b)} Adversarial example computed as described above for $\varepsilon=10$. \textbf{(c)} Same as in (b) but noise is restricted to person pixels only. \textbf{(d)} Adversarial noise. \textbf{(e)} Adversarial noise restricted to person pixels. \textbf{(f)} The adversarial target. \textbf{(g)} Network prediction on adversarial example. \textbf{(h)} Network prediction on adv. example for restricted noise. \textbf{(i)} Network prediction on the original image. \textbf{(j)} Classification difference between (i) and (g). \textbf{(k)} Classification difference between (i) and (h).
% \textbf{(k)} The adversarial target. \textbf{(l)} Classification difference between (k) and (f). \textbf{(m)} Classification difference between (k) and (g).
}
\label{fig:example}
\end{figure}

For an arbitrary given target label $y^{\text{target}}$ adapting \citep{kurakin_adversarial_2016}, the least-likely method iteratively computes an adversarial perturbation $\xi^{(n)}$ as:

\begin{align*}
\xi^{(0)}=0, \ \ \ \xi^{(n+1)}=\text{Clip}_{\varepsilon}\left\{\xi^{(n)}-\alpha\text{sgn}\left(\nabla_{x}J_{\text{cls}}\left(f_{\theta}(x + \xi^{(n)}), y^{\text{target}} \right)\right)\right\}
\end{align*}

Here $\varepsilon$ denotes the maximum $l^{\infty}$-norm of $\xi$ and $\alpha$ the single step-size of one iteration. We set $\alpha=1$, hence changing each pixel maximally by $1$ for each iteration. As suggested in \citet{kurakin_adversarial_2016}, the number of iterations was set to $n=\text{min}(\varepsilon+4, 1.25\varepsilon)$ and adversarial perturbations were evaluated for different $\varepsilon$. The function $\text{Clip}_{\varepsilon}(\xi)$ clips all values of $\xi$ into the intervall $[-\varepsilon, \varepsilon]$.

\paragraph{Definition of adversarial target for semantic segmentation}
\label{paragraph:Definition of adversarial target for semantic segmentation}

In our approach the adversarial target covers the entire image and all pixels of a specific class $c$ are changed towards other classes. To look more natural, we suggest to choose the target class $y^{\text{target}}_{i}$ for all to be fooled pixels $i\in \{j:y^{\text{pred}}_{j}=c\}$ by replacing them with the class of their nearest neighbor that is different from $c$. For all pixels $i$ not of class $c$ in the original network prediction, we set $y^{\text{target}}_{i}=y^{\text{pred}}_{i}$. We measure the effect of adversarial examples by the percentage of pixels of the chosen class that were changed and the percentage of background pixels that were preserved. Because we want to fool the classifier relative to its actual prediction and not the ground-truth, we generate the target label $y^{\text{target}}$ on the basis of the network's actual prediction on the unaltered image.

\section{Results \& Discussion}
\label{section:Results}

Experiments were run with the fully convolutional network architecture introduced by \citet{long_fully_2015} for the VGG16 model. We trained a version of the FCN8 achieving an intersection over union of $55.5\%$. Network training and evaluation as well as generation and evaluation of adversarial examples was done on the downscaled ($1024\times 512$) Cityscapes dataset \citep{cityscapes_2016}.

%\paragraph{Fooling pixels of the same class}
%\label{paragraph:Fooling pixels of the same class}

For the task of hiding all pixels of the same class, we chose to hide all person pixels of the Cityscapes validation dataset. As described above, adversarial target labels $y^{\text{target}}$ were created by first fixating all non-person pixels and second by replacing person pixels with their nearest-neighbor non-person class (see example in Fig.\ref{fig:example} original image (i) and target label (f)). These labels were created for all network predictions on Cityscapes validation images and then used to compute adversarial variations from the images as described above. We used different values for $\varepsilon$ investigating the effect for different magnitudes of adversarial perturbations. In Fig. \ref{fig:example} a representative example is shown, we see that it is possible to fool the deep network into almost not recognising any person pixels while predicting the substituted background and preserving the background outside the original persons nearly perfectly. We also see that the adversarial perturbation is hard to detect.

\begin{figure}[tb]
\begin{center}
\begin{tabular}{cc}
%adv. noise on entire image & adv. noise restricted to persons \\
\includegraphics[width=.49\linewidth]{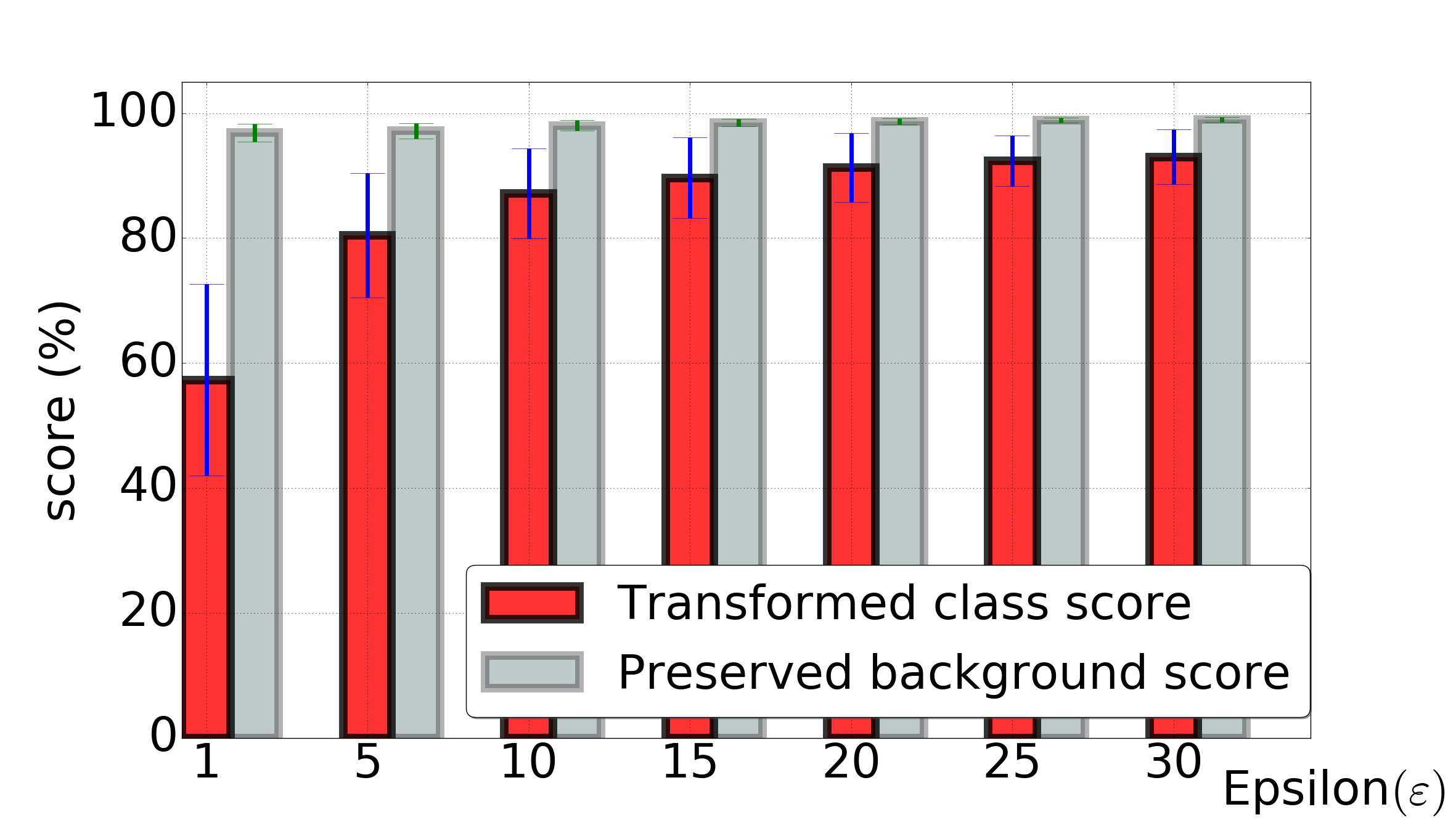} &
\includegraphics[width=.49\linewidth]{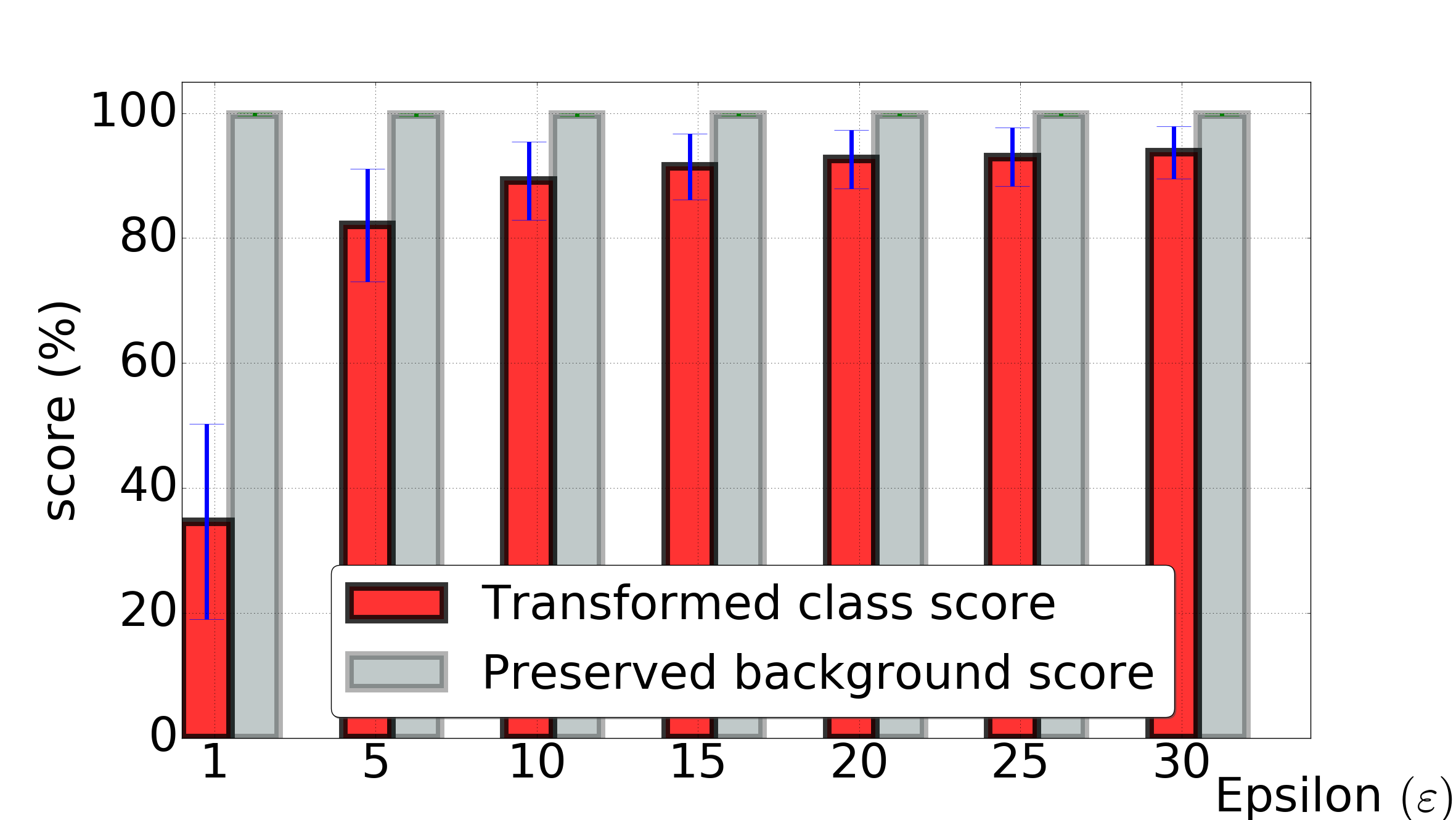} \\
\end{tabular}
\caption{Mean and standard deviation over Cityscapes validation dataset for percentages of preserved (light grey) background and fooled (dark red) person pixels for different $\varepsilon$ and for noise applied to the entire image (left) and restricted to person pixels (right).}
\label{fig:adveval}
\end{center}
\end{figure}

Finally we restricted the adversarial perturbations computed above and applied them only on person class pixels of the network's prediction on the original image. The right column of Fig. \ref{fig:example} shows an example for this case. We see that a majority of person pixels can be cloaked while preserving the background. This indicates that especially the noise on the humans is important to hide them.

Fig. \ref{fig:adveval} shows the mean and standard deviation for percentages of person pixels changed and background pixels preserved for different $\varepsilon$. For the left plot in Fig. \ref{fig:adveval} adversarial perturbations were applied to the entire image, for the right only on person class pixels. In the first case, we see that for sufficiently low $\varepsilon$ of about $10$, over $85\%$ of person pixels could be hidden while over $97\%$ of background pixels were preserved. Considering the case adversarial noise was restricted to person pixels (right plot of Fig. \ref{fig:adveval}), we see that the background is preserved even for smaller $\varepsilon$ while the number of cloaked person pixels decreases for small $\varepsilon$ but recovers for larger values of $\varepsilon$.

\section{Conclusion}
\label{section:Conclusion}

We adapted the concept of adversarial examples to the task of semantic segmentation and showed that existing approaches to generate adversarial examples for classification can be easily transferred to this task. We showed that there exist imperceptible adversarial perturbations that cloak almost all pixels of a target class while leaving the other classes across the image nearly unchanged. Many open topics remain, such as: usage of more performant networks, comparison of different network architectures, more sophisticated methods to measure the effectiveness of adversarial examples for semantic segmentation, can these adversarial examples be applied in the physical world?

\paragraph{Acknowledgments}

We thank the developers of Theano \citep{the_theano_development_team_theano:_2016} and keras (\url{https://keras.io}).

\bibliography{references}

\begin{thebibliography}{7}
\providecommand{\natexlab}[1]{#1}
\providecommand{\url}[1]{\texttt{#1}}
\expandafter\ifx\csname urlstyle\endcsname\relax
  \providecommand{\doi}[1]{doi: #1}\else
  \providecommand{\doi}{doi: \begingroup \urlstyle{rm}\Url}\fi

\bibitem[Cordts et~al.(2016)Cordts, Omran, Ramos, Rehfeld, Enzweiler, Benenson,
  Franke, Roth, and Schiele]{cityscapes_2016}
Marius Cordts, Mohamed Omran, Sebastian Ramos, Timo Rehfeld, Markus Enzweiler,
  Rodrigo Benenson, Uwe Franke, Stefan Roth, and Bernt Schiele.
\newblock The cityscapes dataset for semantic urban scene understanding.
\newblock In \emph{Proc. of the {IEEE} {Conference} on {Computer} {Vision} and
  {Pattern} {Recognition} (CVPR)}, 2016.

\bibitem[Goodfellow et~al.(2015)Goodfellow, Shlens, and
  Szegedy]{goodfellow_explaining_2015}
Ian~J. Goodfellow, Jonathon Shlens, and Christian Szegedy.
\newblock Explaining and {Harnessing} {Adversarial} {Examples}.
\newblock In \emph{{International} {Conference} on {Learning} {Representations}
  (ICLR)}, 2015.

\bibitem[Kurakin et~al.(2016)Kurakin, Goodfellow, and
  Bengio]{kurakin_adversarial_2016}
Alexey Kurakin, Ian Goodfellow, and Samy Bengio.
\newblock Adversarial examples in the physical world.
\newblock \emph{arXiv:1607.02533}, July 2016.

\bibitem[Liu et~al.(2015)Liu, Li, Luo, Loy, and Tang]{parsing_2015}
Ziwei Liu, Xiaoxiao Li, Ping Luo, Chen~Change Loy, and Xiaoou Tang.
\newblock Semantic image segmentation via deep parsing network.
\newblock In \emph{The {IEEE} {International} {Conference} on {Computer}
  {Vision} (ICCV)}, 2015.

\bibitem[Long et~al.(2015)Long, Shelhamer, and Darrell]{long_fully_2015}
Jonathan Long, Evan Shelhamer, and Trevor Darrell.
\newblock Fully convolutional networks for semantic segmentation.
\newblock In \emph{The {IEEE} {Conference} on {Computer} {Vision} and {Pattern}
  {Recognition} (CVPR)}, 2015.

\bibitem[Szegedy et~al.(2014)Szegedy, Zaremba, Sutskever, Bruna, Erhan,
  Goodfellow, and Fergus]{szegedy_intriguing_2013}
Christian Szegedy, Wojciech Zaremba, Ilya Sutskever, Joan Bruna, Dumitru Erhan,
  Ian Goodfellow, and Rob Fergus.
\newblock Intriguing properties of neural networks.
\newblock In \emph{{International} {Conference} on {Learning} {Representations}
  (ICLR)}, 2014.

\bibitem[{The Theano Development
  Team}(2016)]{the_theano_development_team_theano:_2016}
{The Theano Development Team}.
\newblock Theano: {A} {Python} framework for fast computation of mathematical
  expressions.
\newblock \emph{arXiv:1605.02688}, May 2016.

\end{thebibliography}
\bibliographystyle{iclr2017_workshop}

\end{document}